\documentclass{article}
\usepackage{dcase2020,amsmath,graphicx,url,times,booktabs, tabularx}
\usepackage{color}
\usepackage{amsfonts}
\usepackage{cite}
\usepackage{url}
\usepackage[redeflists]{IEEEtrantools}
\newcommand{\iec}{i.\,e.,\,}

\newcommand{\Sect}[1]{Section~\ref{#1}}
\newcommand{\Table}[1]{Table~\ref{#1}}

\title{Low-Complexity Models for Acoustic Scene Classification Based on Receptive Field Regularization and Frequency Damping}
\name{Khaled Koutini$^{1}$,
      Florian Henkel$^{1}$,
      Hamid Eghbal-zadeh$^{1,2}$, 
      Gerhard Widmer$^{1, 2}$
      }
\address{$^1$Institute of Computational Perception (CP-JKU) \& $^2$LIT Artificial Intelligence Lab,\\         
        Johannes Kepler University Linz, Austria\\
        khaled.koutini@jku.at\\ 
 }

\begin{document}
\bstctlcite{IEEEexample:BSTcontrol}
\ninept
\maketitle

\begin{sloppy}

\begin{abstract}

Deep Neural Networks are known to be very demanding in terms of computing and memory requirements. 
Due to the ever increasing use of embedded systems and mobile devices with a limited resource budget, designing  low-complexity models without sacrificing too much of their
predictive performance gained great importance.
In this work, we investigate and compare several well-known methods to reduce the number of parameters in neural networks. 
We further put these into the context of a recent study on the effect of the Receptive Field (RF) on a model's performance, and empirically show that we can achieve high-performing low-complexity models by applying specific restrictions on the RFs, in combination with parameter reduction methods.
Additionally, we propose a filter-damping technique for regularizing the RF of models, without altering their architecture and changing their parameter counts. We will show that incorporating this technique improves the performance in various low-complexity settings such as pruning and decomposed convolution.
Using our proposed filter damping, we achieved the 1st rank at the DCASE-2020 Challenge in the task of Low-Complexity Acoustic Scene Classification.\footnote{Code available at: \url{https://github.com/kkoutini/cpjku_dcase20}}
% Further, we propose a speed-up modification to the decomposition approaches, resulting in faster decomposition, without performance degradation.
\end{abstract}

\begin{keywords}
 low-complexity, acoustic scene classification, receptive-field regularization, pruning, network decomposition
\end{keywords}

\section{Introduction}
\label{sec:intro}

The recent advances in machine learning have been mainly due to the unprecedented successes of deep neural networks with millions or even billions of trainable parameters that can learn from a large amount of data  and solve complex problems.
Although in deep learning, the main attention has been on reaching the highest performance, as these models continue to expand their applications from research into industry, the memory efficiency, energy consumption, and latency of these models become more and more important.

To address these problems, a new line of research has been established to design low-complexity neural networks that are capable of reaching the performance of the large models, while having many orders of magnitude fewer parameters.
In this area, three main approaches have been followed in the literature.
A first approach is \textit{knowledge distillation}~\cite{hinton2015distilling}, which trains a smaller network known as the student, by using the output of a bigger network (the teacher) as the training signal.
Second are methods that focus on proposing efficient neural \textit{architectures} that by design have fewer parameters~\cite{howard2017mobilenets,TanL19effcientnet}.
For example, strategies such as depth-wise separable convolutions~\cite{HowardZCKWWAA17mobile} or careful tuning of the width and depth of the networks~\cite{TanL19effcientnet} aim at producing efficient networks with lower complexity. 
Another approach to designing low-complexity architectures is to use \textit{decomposition layers};
given the fact that neural network parameters are mostly represented with high-dimensional weight tensors, several approaches have been developed to decompose these into smaller tensors to reduce the computational load, without sacrificing the model's performance. 
In~\cite{LebedevGROL14_CNNSVD_ICLR, aderbergVZ14LowRank,KimPYCYS15_TuckerCompression_ICLR}, the authors propose different methods to decompose a single convolutional layer into multiple smaller ones, resulting in more compact models with computational speedup. 
Similarly, \cite{SzegedyLJSRAEVR15_Inception_CVPR, SzegedyVISW16_RethinkingInception_CVPR, heDeepResidualLearning2016} investigate the use of $1 \times 1$ convolutions to reduce the number of channels before applying more expensive operations with larger filter sizes.
A third class of approaches aims at starting with a large high-performance model, to then remove a large part of the parameters by network pruning~\cite{lecun1990optimal,janowsky1989pruning,frankle2018lottery,han2015learningboth}.
\textit{Parameter Pruning} is the process of compressing a neural network by zeroing out some of its parameters. Some methods focus on removing the weights of pre-trained models~\cite{lecun89optimalbraindamagae,MozerS88Skeletonization,janowsky1989pruning}, while more recent approaches incorporate iterative training, by pruning and resetting only the non-pruned weights~\cite{frankle2018lottery}, or pruning without the use of any training data~\cite{Tanaka20synaptic}.
We will focus our analysis on the width and depth restriction methods (Section~\ref{ssec:wd_rest}), on decomposed CNNs (Section~\ref{ssec:decomp}), and  on parameter pruning (Section~\ref{ssec:pruning}), which have gained more popularity and interest among the scientific community, and leave aside the Knowledge Distillation approaches due to their high complexity and slow nature.

% \gerhard{How does this paragraph connect to the previous one?}
% %\hamid{As with the success of Deep Neural Networks (DNNs) in fields such as machine vision, DNNs have also advanced the area of machine listening, and computational auditory scene analysis.}
% 

Recently CNNs have been successfully used for end-to-end Acoustic Scene Classification (ASC)~\cite{Koutini2019Receptive,Chen2019dcase1a,koutinifaresnet2019,Suh2020task1a}, outperforming previous approaches and setting new state-of-the-art.
Recent studies on ASC with CNNs have revealed that regularizing the RF of CNNs can significantly improve their generalisation~\cite{Koutinitrrfcnns2019,koutinifaresnet2019,koutini2019emotion,Suh2020task1a,Koutini2020dcasesubmission}.
Further, authors in~\cite{Koutini2019Receptive} provide a systematic way for controlling the receptive-field of CNNs by adapting the architectural design of the networks.
Although such regularizations improve generalisation, they affect the architectural design and the number of parameters used in a model.
Hence, finding methods that can achieve high performance by following the RF-regularization principles, while having minimal complexity becomes a challenging and important task.

In this paper, we aim at connecting the concept of RF-regularization, with the low-complexity CNNs, and investigate the relationship between the RF size, architecture complexity, and generalisation performance of CNNs for the task of ASC.
To this end, we analyse the performance of different low-complexity ASC methods, under various RFs, 
and show how RF affects the performance in low-complexity settings.
% the 
% pros and cons of the current RF-regularization approaches used in the low-complexity settings.
% 
We empirically evaluate different approaches to low-complexity ASC, and analyse their performance under various maximum receptive fields to reveal the connection between generalisation in a low-complexity setup, and the RF of the models.
Further, we propose a novel RF-regularization technique called ``Damping" which regularizes the RF of any CNN, without a need to alter the topological design of the network.
% , making it a good candidate for using RF-regularization under low-complexity settings.
We show that our Damping RF regularization achieves the best performance with both pruning and decomposition architectures, and hence is a suitable approach for improving generalisation of models in low-complexity settings.

\section{Architectures}
\label{sec:arch}

Previous work has shown the success of RF-regularized CNNs in various acoustic tasks~\cite{Koutinitrrfcnns2019,koutini2019emotion}. 
Therefore, we base our work on the RF-regularized ResNet architecture  introduced in~\cite{Koutini2019Receptive}. 
Furthermore, we introduce a new technique for further restricting the effective receptive field of the network and provide empirical evidence on its success in ASC.

\subsection{Baseline ResNet Architecture: \emph{CP-ResNet}}
\label{ssec:cpresnet}
% We base our experiments on the RF-regularized ResNet architecture~\cite{Koutini2019Receptive}. 
The details of the RF-regularized ResNet architecture are explained in~\cite{koutinifaresnet2019}, where the authors introduce the hyperparameter $\rho$ in the architecture design, 
such that the RF of the architecture can be changed by varying $\rho$.
Since previous work~\cite{Koutini2019Receptive,Koutinitrrfcnns2019,Koutini2020dcasesubmission} showed the optimal range of the RF for different ASC datasets to be approximately between $75$ and $150$ (for the input spectograms explained in Section~\ref{sec:exp}), we restrict our experiments to $\rho$ values in range $3$-$12$. 
Furthermore, we remove the tailing $1 \times 1$ convolutional layers by removing the last 5 residual blocks,\footnote{The residual blocks from 8 to 12 as explained in Table 1 of~\cite{koutinifaresnet2019}} in order to reduce the number of parameters in the baseline. 
As shown in architecture RN1 in~\cite{Koutini2019Receptive}, these tailing layers have a minor effect on the performance of the model.
We refer to this RF-regularized ResNet architecture as \emph{CP-ResNet} throughout this paper.

\subsection{Frequency Damping: \emph{Damped CP-ResNet }}
\label{sec:freq_damp}

Previous work~\cite{Koutini2019Receptive} has shown that restricting the RF of deep CNNs, especially over the frequency dimension, results in better generalization on different ASC datasets.
While \cite{Koutini2019Receptive} introduces a method for systematically tuning the RF of CNNs, the proposed approach requires changes to the architecture, which as a result changes the number of parameters.
To address these drawbacks, we propose a novel method to restrict the \textit{Effective Receptive Field (ERF)} of CNNs -- the part of the RF that has the most influence on the output activation, as detailed in~\cite{luoUnderstandingEffectiveReceptive2016,Koutini2019Receptive} -- by \textit{damping} the convolutional filter weights over the frequency dimensions.
Each convolutional neuron has a limited receptive field of its layer input.
% The damping function reduces the influence of a neuron's input, 
% as the neuron gets further from the center of the input.
Damping works in such a way that the further the input is from the center of a neuron's RF, the less influence it will have on that neuron.
% We reduce the influence of a neuron input, the further away the input is from the center of the neuron's receptive field. 

In practice, we damp the filters of a convolutional layer by applying an element-wise multiplication between the convolution filter weights and a non-trainable constant matrix $C \in \mathbb{R}^{T\times F}$ (damping matrix). 
The damping matrix matches the spatial shape of the filters. 
It decays linearly away from the center, so that the outermost elements of the filter over the frequency dimension have a smaller influence on the activation.

The resulting network is called \emph{damped CNN}, where every convolution  operation $O_n  = W_n * Z_{n-1}   + B_n$ is replaced by $O_n  = (W_n \odot C_n) * Z_{n-1} + B_n$, $*$ is the convolution operator, $\odot$ is the element-wise multiplication operator, $Z_{n-1}$ is the output of the previous layer, $W_n$ is  the filter trainable weight, and $B_n$ is the bias. 
The matrix has a value of 1 in the center, and decays linearly to reach a value $\lambda$; we used $\lambda=0.1$ in our experiments.  
% This simple method can be seen as an inductive bias encoded in the network architecture to encourage the model to make the classification decision based on the information in the middle of its receptive field. 
This approach has shown empirical improvement over \emph{CP-ResNet} in different ASC datasets. We refer to this architecture as \emph{Damp} throughout this paper.

\section{Model Complexity reduction approaches}
\label{sec:approaches}
In this section, we investigate 3 different approaches to reduce the number of parameters of CNNs. 
We follow the principle of reducing the number of parameters, while keeping the final receptive field of the network constant, which allows the comparison with the baseline models in each receptive field setting.
 
\subsection{Width and Depth restriction}
\label{ssec:wd_rest}

Reducing the width (number of channels) and depth (number of layers) of CNNs is a simple technique to reduce the network size. 
Tan and Le~\cite{TanL19effcientnet} showed that after an optimal width of the network is reached, increasing the width further will result in only a minimal performance gain at best. 
The width of the network has quadratic influence on the number of weights, while the depth has a lower effect as the number of parameters grows linearly with the number of layers. 
Based on this fact, and the goal of our experiments that is comparing different RF setups, we focus our efforts on different network widths.

As explained in Section~\ref{ssec:cpresnet}, we remove the tailing residual blocks with $1 \times 1$  convolutions since we are only studying networks with $\rho$ values in range $3- 12$. Therefore, we decrease the depth and the number of parameters from the baseline network in~\cite{Koutini2019Receptive,koutinifaresnet2019}, with only a minimal performance impact. 
However, removing these tailing layers with $1 \times 1$ convolutions result in a significant reduction of the number of parameters in the baseline (reported in~\cite{Koutini2020dcasesubmission} with $\rho=7$)  from $3956K$ to $1715K$.

We change the width of our CNNs (Section~\ref{sec:arch}) by changing the number of channels in initial layers from 128 (in the baseline architecture) to $64$ and $32$. 
This results in reducing the baseline parameter count from $1715K$ to $431K$ and $109K$ respectively.
 %$1715604$ to $431828$ and $109428$ respectively.

\subsection{Decomposed Convolutions}
\label{ssec:decomp}

Inspired by the use of singular-value-decomposition (SVD) for convolutional neural networks~\cite{KimPYCYS15_TuckerCompression_ICLR}, %\cite{LebedevGROL14_CNNSVD_ICLR},
we propose to directly train decomposed convolutional layers instead of decomposing and then fine-tuning a pretrained model, as done in~\cite{KimPYCYS15_TuckerCompression_ICLR} . 
Given a regular convolutional layer with dimensionality 
\begin{equation}
    C_{in} \times C_{out} \times k \times k,
\end{equation}
with $C_{in}$ and $C_{out}$ being the number of input and output filters respectively, and $k$ being the kernel size. 
Such a layer can be decomposed into three convolutional layers using a compression factor $Z$:
\begin{equation}
\begin{split}
    C_{in} \times (C_{out}/Z) \times 1 \times 1\\
    (C_{out}/Z) \times (C_{out}/Z) \times k \times k\\
    (C_{out}/Z) \times C_{out} \times 1\times 1
\end{split}
\end{equation}

For example, a $128 \times 128 \times 3 \times 3$ convolution has $147456$ parameters (neglecting the bias). 
Using a compression factor $Z=4$, we get three convolutions $128 \times 32 \times 1 \times 1$,  $32 \times 32 \times 3\times 3$ and $32 \times 128 \times 1 \times 1$, resulting in 17408 parameters. 
%In this way, we construct a model with less than 20000 parameters which achieves more than $95.8\%$ accuracy on the development set.
Similar approaches to decompose convolutions for the purpose of parameter reduction are explored in~\cite{SzegedyLJSRAEVR15_Inception_CVPR, SzegedyVISW16_RethinkingInception_CVPR, heDeepResidualLearning2016}. 
Our proposed decomposition shares the same structure as the ''bottleneck'' building block described in~\cite{heDeepResidualLearning2016}, but is different from them as we do not use non-linearities and batch-normalization within the decomposed block, which proved to yield better performance.%\flo{should we add one sentence saying that not using a non-linearity turned out to work better, or we just skip that?}

\subsection{Parameter Pruning}
\label{ssec:pruning}

Pruning is well studied in neural network literature and several methods and adaptation have been proposed~\cite{frankle2018lottery,han2015learningboth,lecun89optimalbraindamagae,MozerS88Skeletonization,janowsky1989pruning,Tanaka20synaptic}.
We use magnitude pruning~\cite{janowsky1989pruning,han2015learningboth} with iteratively increasing the pruning ratio until reaching the desired number of parameters. 

% Since previous work~\cite{Tanaka20synaptic} shows simple magnitude pruning achieves similar performances in our required compression rate. 
Authors in~\cite{Tanaka20synaptic} show that pruning approaches perform very similarly, if the compression-ratio is up to 1\%.
Since in our low-complexity setting, we are not targeting compression-rates lower than 1\%, we choose to use the more simple magnitude-pruning approach.
% 
% Starting from zero, we continually increase the pruning-rate such that the total number of non-pruned parameters follows an exponential decay function from the baseline number of parameters in the first epoch to the desired number of parameters in epoch 150. 
We ramp up the number of pruned parameters with an exponential decay from 0 to the final desired number in 100 epochs, to get models with 250K, 300k, 400k, and 500k parameters.
This allows us to remove more weights at the beginning of the training, and fewer weights in later stages as the model converges.

\begin{figure}[t]
%fa_analysis-2d_dcase_freq_rime.ipynb
\centering
\includegraphics[width=3.5in]{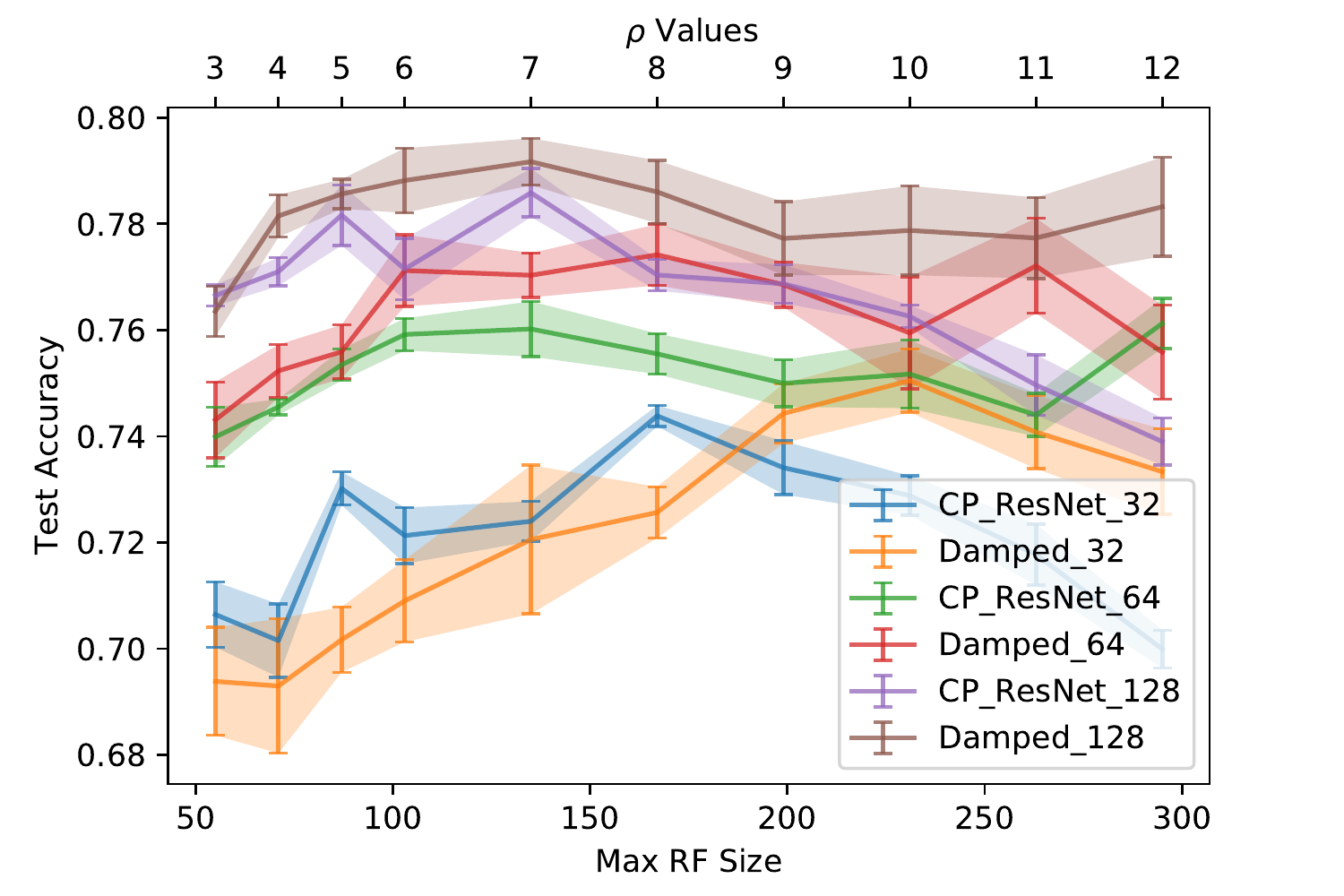}
\caption{Testing accuracy of CP-ResNet on DCASE'18 with/without filter damping on the frequency dimension using different CNN width setups as explained in Section~\ref{ssec:wd_rest}. }
\label{fig:dcase18:damped:vs:base}
\end{figure}

\section{Experimental Setup}
\label{sec:exp}
%\flo{add brief introductory paragraph}
%\subsection{Datasets}
We evaluate our proposed approaches introduced in \Sect{sec:approaches} on two acoustic scene classification (ASC) datasets.

\textbf{DCASE'20 Low Complexity ASC}~\cite{Heittola2020}:
The dataset contains 40 hours of recordings from 12 European cities in 10 different acoustic scenes that are summarized into three categories, \emph{indoor}, \emph{outdoor} and \emph{transportation}. We follow the training/test split provided by the task organizers where recordings from one city are not seen during training. 

\textbf{DCASE'18 ASC}~\cite{MesarosDCASE2018T1}:
The dataset comprises recordings from six European cities with similar acoustic scenes as the DCASE'20 dataset. %  In contrast to DCASE'20, the goal here is to predict one of 10 specific scene labels. Due to the higher number of classes this dateset provides a more difficult challenge allowing us to better assess the performance of our systems. 
We choose this dataset because it has 10 classes which makes it a more challenging task compared to DCASE'20, therefore we can better differentiate the performance between different approaches with the 500KB limit (and more) that was used in the DCASE'20 challenge.
This dataset has a relatively smaller size resulting in shorter training time. Therefore, we use DCASE'18 for our detailed analysis.

% \subsection{Data Preparation and Training}
% \label{ssec:preparation}

For feature extraction, we compute perceptually weighted Mel-spectrograms from the provided 10 second audio snippets which we down-sample to 22.05kHz similar to~\cite{Koutini2019Receptive}. 
Each input channel of the stereo audio is processed independently, normalized using the training set statistics and provided as a two-channel-spectrogram input to the CNN. We use the same experiments and training setup as explained in ~\cite{koutinifaresnet2019}.

% For model training, we use the Adam optimizer \cite{kingmaAdamMethodStochastic2014} with a custom learning rate scheduler that gradually increases the learning rate from $1e^{-6}$ to $1e^{-4}$ over the course of 10 warm-up epochs. From epoch 10 until 50 we keep a constant learning rate of  $1e^{-4}$ and start to linearly decrease it back to  $1e^{-6}$ from epoch 50 until epoch 220. Finally, we train for another 30 epochs with the minimum learning rate $1e^{-6}$.

% As additional data augmentation, we apply \emph{Mix-up} \cite{zhangMixupEmpiricalRisk2017}, which linearly combines two input samples and their targets to improve model generalization.
% %\flo{Should we mention SWA as well?}

\section{Results}

\begin{figure}[t]
%fa_analysis-2d_dcase_freq_rime.ipynb
\centering
\includegraphics[width=3.5in]{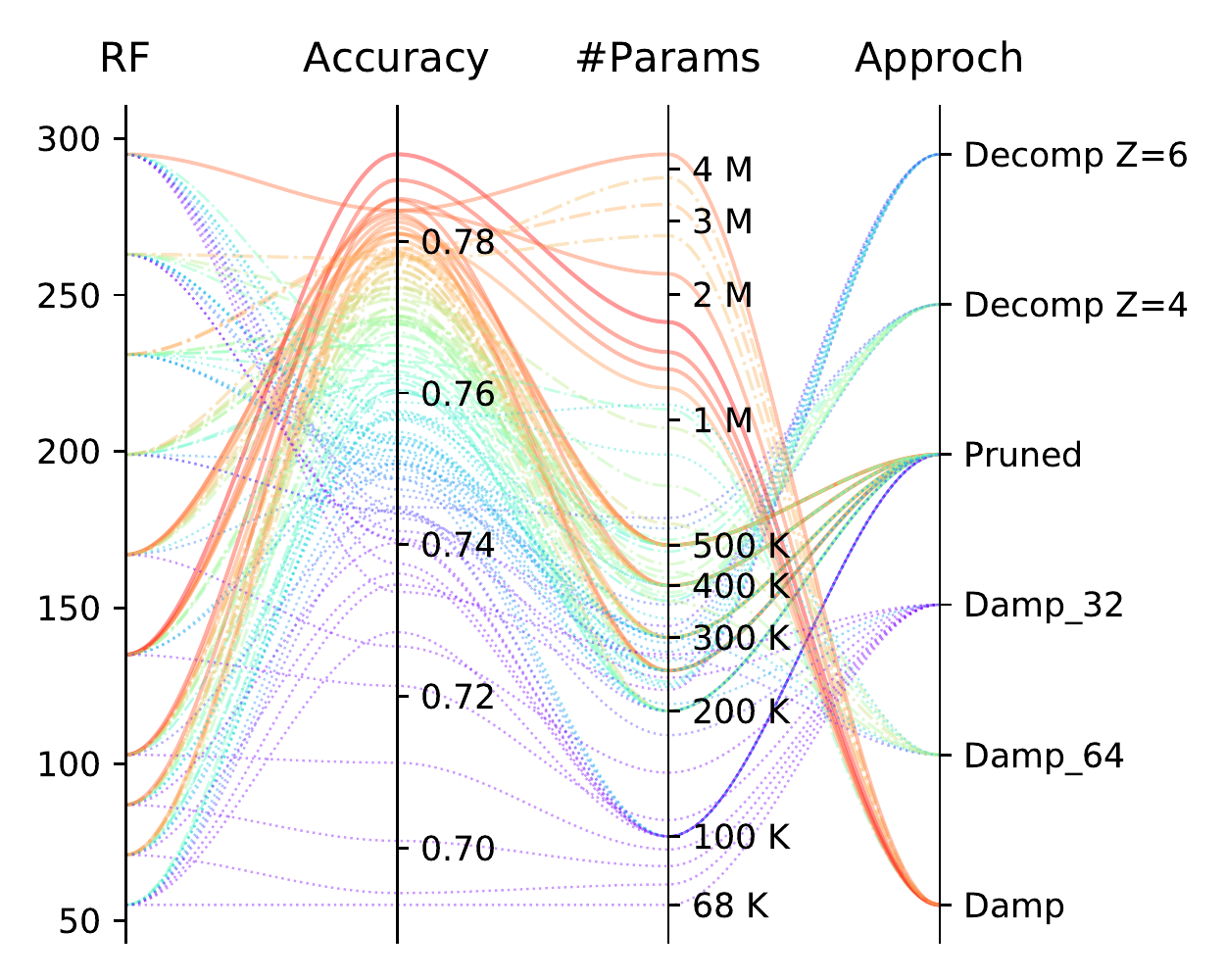}
\caption{Comparison of different complexity reduction methods in different receptive field setups on DCASE'18. All methods apply frequency-damping. The line color represents the accuracy.}
\label{fig:dcase18:damped:overview}
\end{figure}

\begin{figure}[t]
%fa_analysis-2d_dcase_freq_rime.ipynb
\centering
\includegraphics[width=3.5in]{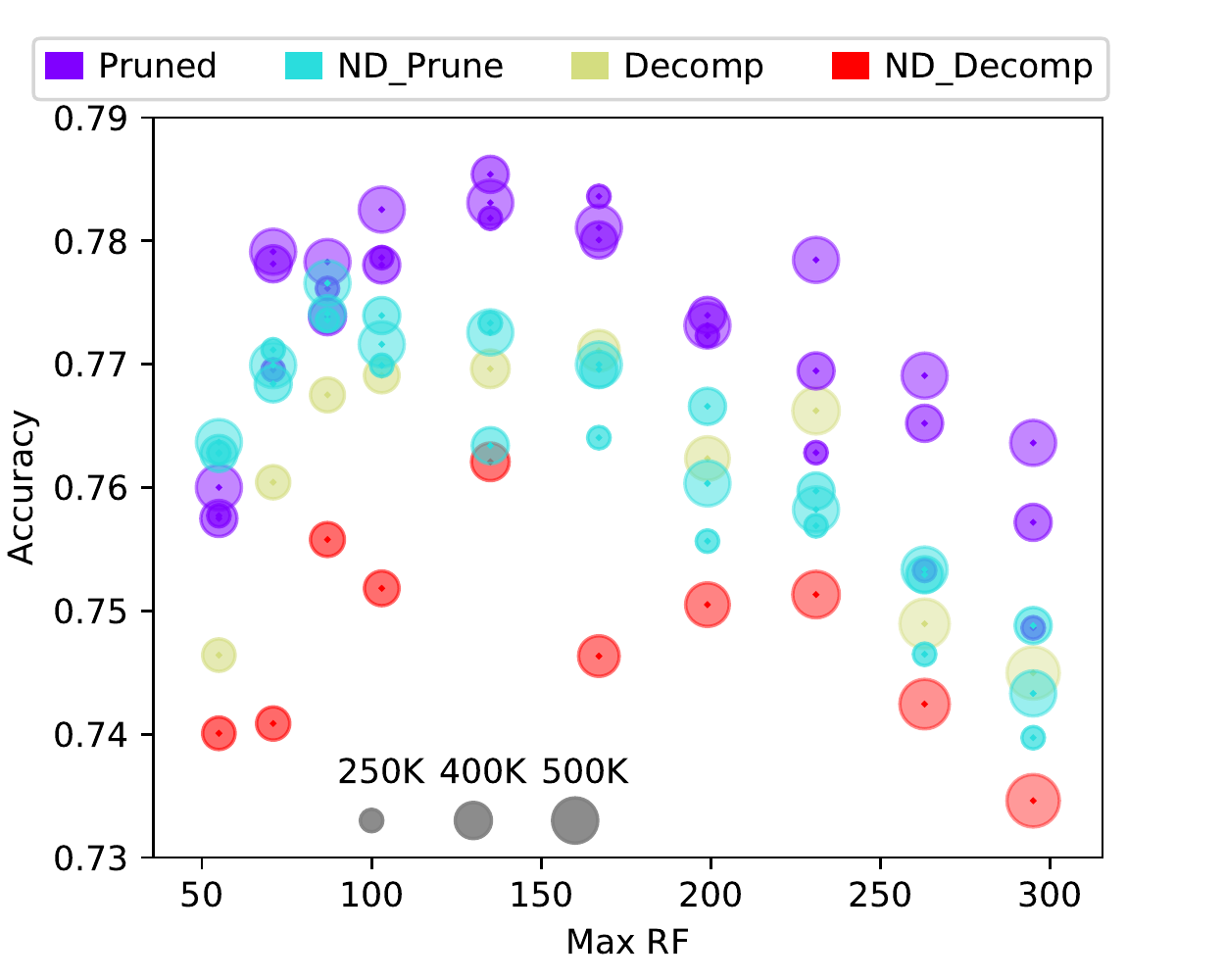}
\caption{Comparing different complexity reduction methods in different receptive field setups with and without damping on DCASE'18. All methods use 128 base channels. Decomp uses a compression factor $Z=4$. ND refers to no damping method being applied to the original Network.
The size of the dots relates to the number of model parameters.}
\label{fig:dcase18:dampedvsnodamped:prune:vs:decomp:disk}
\end{figure}

\begin{figure}[t]
%fa_analysis-2d_dcase_freq_rime.ipynb
\centering
\includegraphics[width=3.5in]{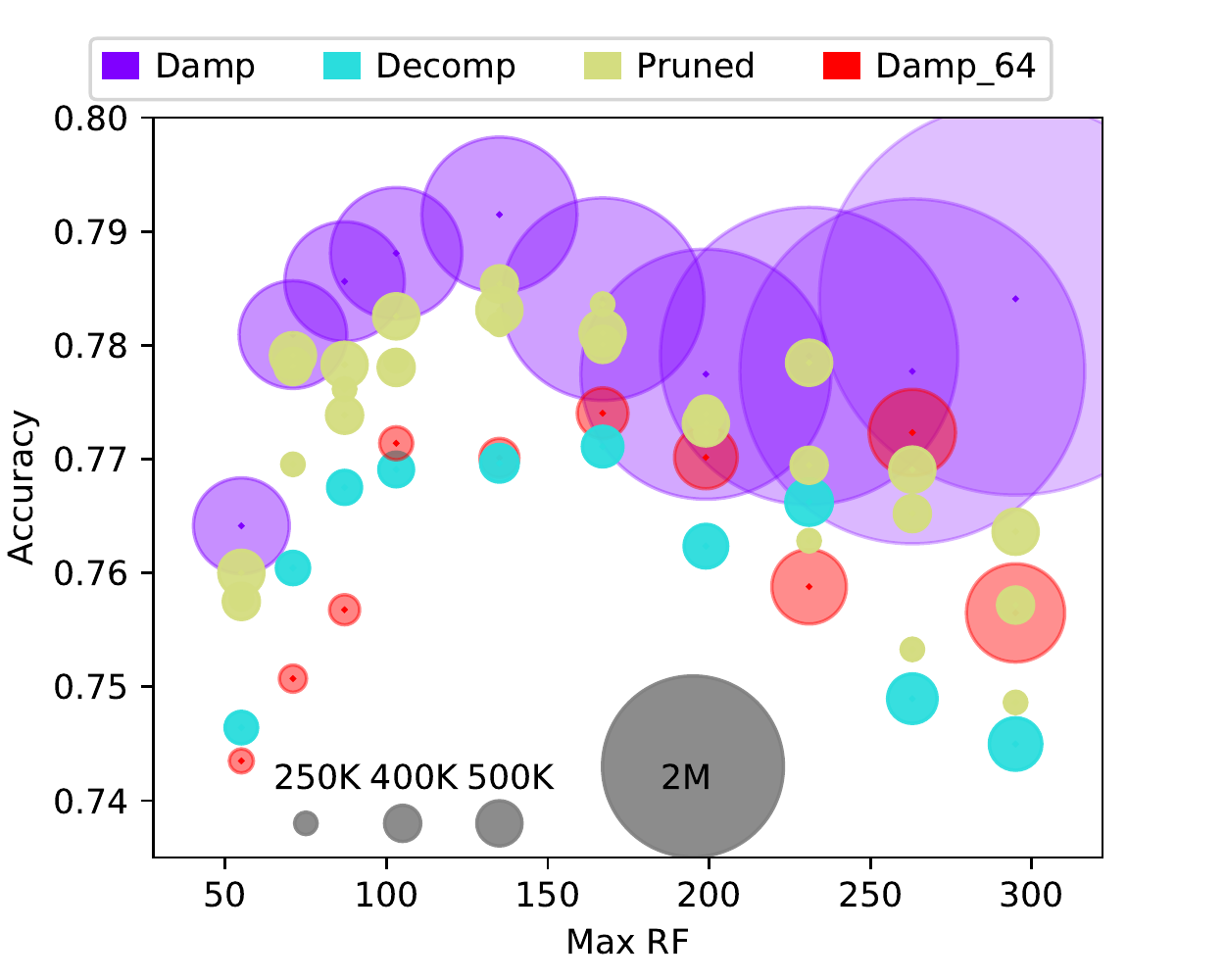}
\caption{Comparing different complexity reduction approaches on the DCASE'18 dataset. All networks are damped. 
The size of the dots relates to the number of model parameters.}
\label{fig:dcase18:damp:prune:vs:decomp}
\end{figure}

% \begin{figure}[ht]
% %fa_analysis-2d_dcase_freq_rime.ipynb
% \centering
% \includegraphics[width=3.5in]{figs/par_prune_vs_decomp_desk2.pdf}
% \caption{Comparing different complexity reduction methods in different receptive field setups. All methods use 128 base channels and Damped.}
% \label{fig:dcase18:prune:vs:decomp:disk2}
% \end{figure}

\Table{tab:results_overview} summarizes the results of our best performing models on the DCASE'20 and DCASE'18 datasets. 
On both datasets we observe an improvement for the damped ResNet architecture, compared to the baseline.
Furthermore, we observe that a simple width reduction, \iec reducing the number of base channels and thus decreasing the overall number of parameters, is not as effective as pruning the weights of a bigger network. 
Following the pruning strategy explained in \Sect{ssec:pruning}, we can achieve models with a significantly lower number of non-zero parameters, with the same performance on DCASE'20 or only minor performance degradation on DCASE'18.
% of less than ... and 1 percentage point for DCASE'20 and DCASE'18, respectively.
%In the following we take a closer look on the results of each parameter reduction approach considering different maximum receptive fields.

\begin{table}[!htbp]
\small
\begin{tabular}{lrrrr}
\toprule
&\multicolumn{2}{c}{\textbf{DCASE'20}}                       & \multicolumn{2}{c}{\textbf{DCASE'18}} \\\midrule
\textbf{Model} & \textbf{Non-Zero} & \textbf{Acc.}         &   \textbf{Non-Zero} & \textbf{Acc.} \\ \toprule
 CP-ResNet        & 1061.6K    & .9726           & 1715.6K    & .7852   \\       %dcase20 rf 55
 Damp.            & 1061.6K    & .9745           &  1715.6K   & .7915                  \\ 
 Damp.-Restricted &   268.6K     & .9722           &  431.8K    & .7701                  \\ 
 Damp.-Pruned     &   250K     &  .9745         &  400K      & .7854                   \\ 
 Damp.-Decomp.    &   361K     & .9700           &  417K    & .7696   \\\toprule
\end{tabular}
\caption{Comparison of the performance of the best models of the discussed approaches on the DCASE`20 and DCASE'18 datasets.  We report the accuracy (Acc.) and the number of non-zero parameters. 
\emph{Damp.} refers to a damped ResNet as introduced in \Sect{sec:freq_damp}, \emph{Damp.-Restriced} is a width restricted ResNet with 64 base channels (cf. \Sect{ssec:wd_rest}), \emph{Damp.-Pruned} is a  pruned ResNet (cf. \Sect{ssec:pruning}) and \emph{Damp.-Decomp.} is a ResNet where all convolutions are replaced with decomposed convolution and a compression factor $Z=4$ (cf. \Sect{ssec:decomp}). The table shows the mean of the accuracy over the last 10 epochs for each model. All DCASE'20 and DCASE'18 models use  $\rho=4$ and $\rho=7$, respectively.
}
\label{tab:results_overview}
\end{table}

\subsection{Frequency Damping Comparison}

To further investigate the effect of \emph{damping}, we compare the performance of several baseline CP-ResNets with and without \emph{damping}, using different maximum RF setups on DCASE'18. Fig.~\ref{fig:dcase18:damped:vs:base} shows that \emph{damping} not only improves the overall performance, but is also more robust to the choice of the maximum RF. %which we attribute to the fact that this method additionally restricts the effective receptive field. 
We also observe that the damped ResNet with 128 base channels (Damped) yields the best overall performance using $\rho=7$ to regularize the RF. 
The model with 64 base channels (Damped\_64) decreases the performance, while significantly reducing the number of parameters (shown in Figure~\ref{fig:dcase18:damped:overview}), and 32 base channels seems to result in an insufficient amount of parameters for this task.
As can be seen in Fig.~\ref{fig:dcase18:dampedvsnodamped:prune:vs:decomp:disk},  Damped CP-ResNets outperforms non-damped  CP-ResNets with the same number of parameters, both in pruning and decomposition cases.

\subsection{Parameter Pruning vs. Network Decomposition} %vs. Width Reduction}

Comparing all parameter reduction methods in Figs.~\ref{fig:dcase18:damped:overview} and ~\ref{fig:dcase18:damp:prune:vs:decomp}, we see that pruning is the closest model to the frequency-damped baseline (Damp) in terms of performance, and achieves the highest accuracy with limited parameters. 
Fig.~\ref{fig:dcase18:damp:prune:vs:decomp} further shows that the pruned model achieves an accuracy close to the baseline, while significantly reducing the number of non-zero parameters. 
% This requires careful tuning, as too little parameters will result in severe performance degradation.
% Using decomposed convolutions can reduce the number of parameters even further. However, the performance degradation is significantly higher than for pruning. 
% As can be seen in \Table{tab:results_overview}, the width-reduction approach performs the worst among the parameter reduction methods.
Using decomposed convolutions, the performance degradation is significantly higher than for pruning. Similarly, simply reducing the width of the network results in a large loss in accuracy.

\section{Conclusion}
In this paper, we analysed various low-complexity CNN-based approaches for ASC, and studied the relationship between the RF and the performance in each approach.
We showed that pruning achieves better performance compared to decomposition and width reduction methods.
We proposed a filter damping technique that can be used in low-complexity settings to regularize the RF of models, without altering the architectures.
We showed that using filter damping improves the performance on all architectures and datasets we evaluated, hence is a simple and effective technique for improving generalisation of models.
% Additionally, we proposed a simplification to pruning techniques that while achieving the best performance in low-complexity setting, is faster compared to its counterparts.
Our results on two datasets for ASC suggests that the proposed filter damping can achieve state-of-the-art performance in low-complexity ASC. 
Using this approach, we achieved the 1st rank at the DCASE-2020 Challenge \textit{Low-Complexity Acoustic Scene Classification} (Task1.b).

\section{ACKNOWLEDGMENT}
\label{sec:ack}
 This work has been supported by (1) the COMET-K2 Center of the Linz Center of Mechatronics (LCM) funded by the Austrian federal government and the federal state of Upper Austria, and (2) the European Research Council (ERC) under the European Union's Horizon 2020 research and innovation program (grant agreement number 670035, project "Con Espressione").
\newpage
\clearpage
\bibliographystyle{IEEEtran}
\bibliography{refs}

% \begin{figure}[ht]
% %fa_analysis-2d_dcase_freq_rime.ipynb
% \centering
% \includegraphics[width=3.5in]{figs/damp_all_desk.pdf}
% \caption{This figure can replace figure 4. }
% \label{fig:dcase18:prune:vs:decomp_disk}
% \end{figure}

\end{sloppy}
\end{document}